\documentclass[letterpaper, 10 pt, conference]{ieeeconf}  

\IEEEoverridecommandlockouts                              

\usepackage{hyperref}
\usepackage{graphics} 
\usepackage{amsmath} 
\usepackage{amssymb}  
\usepackage{mathtools}
\usepackage[capitalise]{cleveref}
\usepackage{algpseudocode}
\usepackage{graphicx}
\usepackage{booktabs}
\usepackage{xcolor}
\usepackage{algorithm}
\usepackage{wrapfig}
\usepackage{makecell}
\usepackage{cite}
\usepackage{subcaption}
\usepackage{oplotsymbl}
\usepackage{amssymb}
\usepackage{wasysym}

\usepackage{cleveref}
\usepackage[all]{custom_format}

\usepackage[textsize=small]{todonotes}

\newcommand\scalemath[2]{\scalebox{#1}{\mbox{\ensuremath{\displaystyle #2}}}}

\title{\LARGE \bf
Symmetry-aware Reinforcement Learning for Robotic Assembly under Partial Observability with a Soft Wrist
}

\author{Hai Nguyen$^{1*,2}$, Tadashi Kozuno$^{1}$, Cristian C. Beltran-Hernandez$^{1}$, Masashi Hamaya$^{1\dagger}$%
\thanks{$^{1}$OMRON SINIC X Corporation, Hongo 5-24-5, Bunkyo-ku, Tokyo, Japan. $^{*}$Work done as an intern at OMRON SINIC X. $^{2}$Northeastern University, Boston, MA 02115, USA. $^{\dagger}$This study is supported by JST ACT-X, Grant Number JPMJAX22AC. Code and videos are available at \url{https://github.com/omron-sinicx/symmetry-aware-pomdp}.}
}

\begin{document}

\maketitle
\thispagestyle{empty}
\pagestyle{empty}

\begin{abstract}
This study tackles the representative yet challenging contact-rich peg-in-hole task of robotic assembly, using a soft wrist that can operate more safely and tolerate lower-frequency control signals than a rigid one. Previous studies often use a fully observable formulation, requiring external setups or estimators for the peg-to-hole pose. In contrast, we use a partially observable formulation and deep reinforcement learning from demonstrations to learn a memory-based agent that acts purely on haptic and proprioceptive signals. Moreover, previous works do not incorporate potential domain symmetry and thus must search for solutions in a bigger space. Instead, we propose to leverage the symmetry for sample efficiency by augmenting the training data and constructing auxiliary losses to force the agent to adhere to the symmetry. Results in simulation with five different symmetric peg shapes show that our proposed agent can be comparable to or even outperform a state-based agent. In particular, the sample efficiency also allows us to learn directly on the real robot within 3 hours.
\end{abstract}

\section{INTRODUCTION}

A peg-in-hole task often abstracts an insertion operation in robotic assembly~\cite{JiangRCIM2022}. This task is widely explored~\cite{JiangRCIM2022} but is still challenging due to the contact-richness and tight clearance between parts.
Most approaches involve rigid robots with force control~\cite{xu2019compare}, sending high-frequency control signals for safe operation. As an alternative, physically soft robots can be used for contact-rich tasks, given their softness can tolerate contacts and low-frequency control signals without sacrificing safety~\cite{GotoSMC1980, azulay2022haptic, brahmbhatt2023zero, HartischArxiv2023}.

Controlling soft robots, however, is challenging due to their soft components' non-linear nature. Hence, learning-based control is beneficial for soft robots. Most previous studies assume a fully observable setting in which the robot's pose is obtained using motion capture systems or vision-based estimators~\cite{yasa2023overview}. However, these methods have limitations, such as setup constraints and potential inaccuracies, particularly in occluded scenarios. In contrast, this work adopts a partially observable Markov decision process (POMDP) approach, where the relative peg-to-hole pose is treated as unknown. The focus is on training a memory-based agent using force/torque (F/T) feedback and proprioceptive information of the robot. However, dealing with POMDPs is more complex than MDPs, often requiring an agent to memorize and actively gather information besides acting purely for rewards~\cite{kaelbling1998planning}. Therefore, we need to find a useful inductive bias to learn more efficiently for this task.

\begin{figure}
    \centering
    \includegraphics[width=0.95\linewidth]{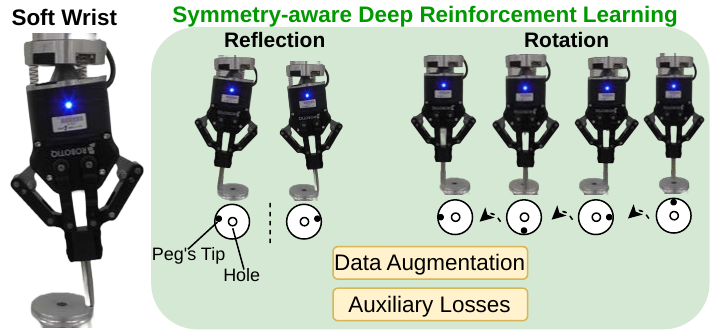}
    \caption{We leverage the domain symmetry to augment data (through reflections and rotations) and to regularize (though auxiliary losses) to learn a symmetry-aware agent.}
    \label{fig:intro}
    \vspace{-15pt}
\end{figure}

This study proposes to leverage the potential symmetry of the task for learning a memory-based actor-critic agent in a sample-efficient manner. Specifically, the peg and the hole (as well as many industrial parts) are often symmetric or highly similar on multiple sides for easy assembly. To motivate our approach,~\cref{fig:intro} depicts six initial peg-to-hole positions, including two reflected and four rotational positions. These positions are equivalent due to the symmetry of a round hole. Consequently, one can transform (rotate or reflect) one trajectory initiated in one position to create valid trajectories in the other positions, indicating the possibility for data augmentation. Moreover, because these trajectories are equivalent, a critic should evaluate them the same. Similarly, an actor making the next decision based on the transformed trajectories should also transform its next actions similarly. To sum up, the following are our contributions:
\begin{itemize}
    \item We leverage the symmetry of the peg-in-hole task for data augmentation and regularizing the actor and the critic through auxiliary losses to reduce the search space for more sample-efficient learning. To the best of our knowledge, enforcing the symmetry this way for POMDPs has not yet been studied in previous works.
    \item We conduct experimental evaluations across five symmetric peg shapes in simulation using MuJoCo~\cite{todorov2012mujoco}. The results show that our proposed agent can even be positively comparable to a state-based agent. Moreover, the learned policy can be generalized from one shape to another. In particular, the sample efficiency allows us to learn the task directly on a real robot using 100 demonstrations within 3 hours.
\end{itemize}

\section{RELATED WORKS}
\subsection{Pose Estimation in Soft Robots}
Motion capture systems are often used to measure soft robots' poses~\cite{yasa2023overview,bruder2019nonlinear,bern2020soft}. Soft robotic assembly scenarios~\cite{hamaya2020learning_a, hamaya2020learning_b} also assumed a full observability setting in which the peg-to-hole pose is measured using an external motion capture setup.
As alternative approaches, embedded bending sensors~\cite{thuruthel2019soft}, air-pressure~\cite{kawase2021pneumatic} or, multi-modal sensors~\cite{navarro2020model, loo2022robust} have been used. Instead, this study uses F/T sensors to obtain the contact states of the robot indirectly. A relevant study uses an F/T sensor to estimate an elastic rod's shape with a model-based approach~\cite{takano2017real}. In contrast, we do not require a model assumption and instead employ a model-free deep reinforcement learning (DRL) approach considering the POMDP setting.

\subsection{Peg-In-Hole with Soft Robots using DRL}

Combining F/T signals and DRL for policy learning in the peg-in-hole task using rigid robots exists in several recent studies such as~\cite{lee2019making, beltran2020variable, ding2019transferable}. For the methods,~\cite{lee2019making} fuses F/T signals with RGB images, while variable compliance control is combined with a model-free approach in~\cite{beltran2020variable}. In contrast,~\cite{ding2019transferable} utilizes a model-based approach by learning a force-torque dynamics model. Closest to our setting is~\cite{azulay2022haptic}, which uses a model-free approach in a POMDP formulation to learn an F/T-based peg-insertion policy using an under-actuated compliant hand. However, the policy relies on pre-defined motion primitives to ease learning and uses DRL to produce residual signals to assist these primitives. In contrast, our agent does not count on any primitives.

\subsection{Symmetry-aware Policy Learning}
Exploiting the domain symmetry can benefit robot learning, especially when samples are often expensive to collect with real robot systems. Under full observability, recent studies such as~\cite{wang2022so, zhu2022sample, lee2023sample} introduced various symmetry-aware RL agents that encode the domain symmetry directly into their network architectures. In particular,~\cite{li2023efficient} instead uses the symmetry to transform the state and feed both the original and transformed states to the actor and the critic, making an actor-critic agent symmetry-aware. In contrast to the fully observable setting, incorporating symmetry is under-explored under partial observability, which is considered more difficult to solve. A recent work in this direction is~\cite{nguyen2023equivariant}, which enforces symmetry by directly constructing every component of an actor-critic agent with \emph{equivariant} (symmetry-aware) neural networks~\cite{e2cnn}. In contrast, we use no specialized neural networks, and instead \emph{indirectly} enforce the symmetry through auxiliary losses and additionally use the domain symmetry for data augmentation, applied to a recurrent Soft Actor-Critic (SAC)~\cite{haarnoja2018soft} agent. Moreover, the way we use auxiliary losses to enforce the symmetry differs from the common practices in the POMDP literature, in which auxiliary tasks are often used to learn better history representations through reconstruction tasks~\cite{nguyen2020belief} or prediction tasks~\cite{baisero2020learning}.

\section{BACKGROUND}
\subsection{Partially Observable Markov Decision Processes}
\begin{figure}[htbp]
    \centering
    \includegraphics[width=1.0\linewidth]{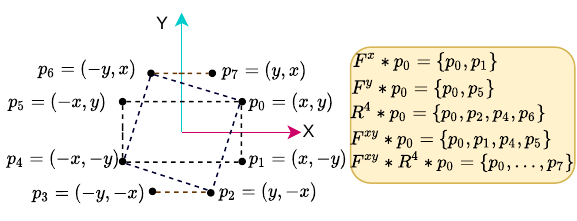}
    \vspace{-15pt}
    \caption{Transforming a 2D point $p_0 = (x, y)$ using X-axis reflection ($F^x$), Y-axis reflection ($F^y$), counter-clockwise rotations of $\{0, \pi/2, \pi, 3\pi/2\}$ around the origin ($R^4$), and their sequential combinations ($F^{xy} = F^x*F^y$ and $F^{xy}*R^4$).}
    \label{fig:viz-augment}
    \vspace{-15pt}
\end{figure}
A POMDP~\cite{astrom1965optimal} is defined by a tuple $(\mathcal{S}, \mathcal{A}, \Omega, T, R, O)$, where $\mathcal{S}$, $\mathcal{A}$, and $\Omega$ are the state space, the action space, and the observation space, respectively. After an action $a$, the state changes from $s$ to $s'$ according to the dynamics function $T(s, a, s')$. The transition emits an observation $o$, decided by the observation function $O(a, s', o)$. For optimal behavior, an agent may need to rely on the entire observable action-observation history $h_t = (o_0, a_0, \ldots, a_{t-1}, o_t)$~\cite{singh1994learning}. Therefore, the goal is to find a history-based policy $\pi(h_t)$ that maximizes the expected discounted return $J = \Exp\left[ \sum_{t=0}^\infty \gamma^t R(s_t, a_t) \right]$, with a discounting factor $\gamma \in [0, 1)$. In this study, the history-based policy is the actor of a recurrent SAC agent. Moreover, instead of learning a state-based critic, a history-based critic $Q(h_t, a_t)$ is learned during training.

\subsection{Notations}

\begin{wrapfigure}[12]{R}{0.45\linewidth}
  \centering
  \includegraphics[width=1.0\linewidth]{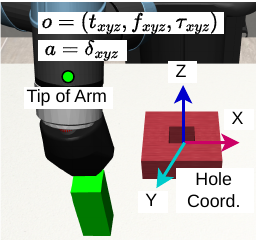}
  \caption{Our sim. setup.}
  \label{fig:coordinates}
\end{wrapfigure}

We use $G$ to denote a set of group elements $g$; each defines a transformation. A special element of $G$ is the identity transformation $e$, which leaves the input unchanged. This work focuses on 2D transformations; therefore, $G$ is limited to planar reflections and rotations. For the planar reflections, we use $F^x = \{ e, r^x \}$ for the X-axis reflection, with $r^x$ defining the reflection. We denote the result of applying $G$ on a 2D point $p_0 = (x, y)$ as $G*p_0$, which results in a new set, whose elements are $g*p_0$ for each $g \in G$. For instance, $F^x*p_0 = \{e*p_0, r^x*p_0\} = \{p_0, p_1 \}$ (see~\cref{fig:viz-augment}). Similarly, $F^y = \{ e, r^{y} \}$ refers to the Y-axis reflection. For convenience, we use $F^{xy}$ to shorthand for $F^x*F^y$, which combines the two transformations sequentially, i.e.,  $F^{xy}*p_0 = \{p_0, p_1, p_4, p_5\}$. For rotational transformations, we use $R^n$ to refer to a set of $n$ planar rotations $\{0, 2\pi/n, 4\pi/n, \dots, 2\pi(n-1)/n\}$ ($e$ corresponds to a rotation of $0$ rad). Specifically, a transformation element $g \in R^n$ will rotate $p_0$ counter-clockwise around the origin. For instance, $R^4*p_0 = \{p_0, p_2, p_4, p_6 \}$. Similar to $F^{xy}$, we can combine $F^{xy}$ and $R^4$, i.e., $F^{xy}*R^4*p_0 = (p_0, \dots, p_7)$. Finally, we use $|G|$ to denote the group size of $G$, e.g., $|R^n| = n$, $|F^{x}| = |F^{y}|=2$, $|F^{xy}|=4$, and $|F^{xy}*R^4| = 8$.

\section{SYMMETRIC PEG-IN-HOLE AS A POMDP}
\label{sect:problem_formulation}
We consider the task of inserting a peg into a \emph{symmetric} hole as a POMDP, in which an observation $o \in \Omega$ includes 3D position $(t_x,t_y,t_z)$ of the tip of the arm, 3-axis torques $(\tau_x,\tau_y,\tau_z)$ and forces $(f_x,f_y,f_z)$ (see~\cref{fig:coordinates}). All these signals are converted to a coordinate defined at the hole's center (the hole coordinate). Like the haptic feedback when humans perform similar tasks, an F/T sensor can implicitly tell the peg-to-hole relative pose~\cite{qiao1993robotic, tang2016autonomous}. In particular, when the peg and the hole are symmetric, the symmetry can manifest through the F/T signals. We use a continuous action $a = (\delta_x, \delta_y, \delta_z)$, indicating the displacements of the arm's tip. Note that rotational adjustments along the XY axes can be made by pushing down (to leverage the robot's softness) and then applying the tip displacements along the XY axes. Furthermore, we assume the peg is already well-aligned to the hole in the rotation around the Z axis. The agent is given a sparse reward when the peg is successfully inserted into the hole. In the fully observable version, a state $s \in \mathcal{S}$ contains the relative pose from the peg to the hole coordinate, denoted as $p_{p2h}$. Finally, we summarize our setting below:
\begin{equation} \label{eq:setting}
\begin{split}
    o &= (t_x, t_y, t_z, f_x, f_y, f_z, \tau_x, \tau_y, \tau_z), \quad a = (\delta_x, \delta_y, \delta_z), \\
    s &= p_{p2h} , \quad r = 1 \text{ only if task accomplished}.
\end{split}
\end{equation}

\section{PROPOSED METHOD}

\begin{figure}[htbp]
    \centering
    \includegraphics[width=1.0\linewidth]{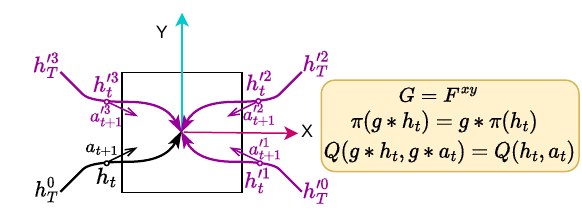}
    \vspace{-10pt}
    \caption{The hole symmetry allows data augmentation on a history $h_T$ through the transformation group $G = F^{xy}$. Moreover, given the current history $h_t$, the Q-function and the policy need to satisfy specific properties for $\forall g \in G$.}
    \label{fig:2d-symmetry}
    \vspace{-10pt}
\end{figure}

In~\cref{fig:2d-symmetry}, we choose a square peg to illustrate the benefits of using domain symmetry. Specifically, assuming we have a 2D trajectory $h^0_T$ (black), projected from some 3D trajectory that successfully inserts the peg into a square hole. Suppose we consider the group transformation $G = F^{xy}$, then we can transform each point in $h^0_T$, resulting in three more equivalent trajectories $(h'^1_T, h'^2_T, h'^3_T)$. This indicates the possibility of augmenting the training data using domain symmetry, which can also apply to failed trajectories. \cref{fig:2d-symmetry} also indicates symmetry properties of the Q-function and the policy if we use an actor-critic agent for this task. Given the current history $h_t$ of the trajectory $h^0_T$, then $Q(h_t, a_t)$ should be equal to $Q(h'^1_t, a'^1_t)$, $Q(h'^2_t, a'^2_t)$, and $Q(h'^3_t, a'^3_t)$, where $a'^i_t$ with $i=1, 2, 3$ are the transformed version of $a_t$ through $G$. In other words, we want the Q-function to be invariant to $G$:
\begin{align}\label{eq:invariant_critic}
    Q(h_t, a_t) = Q(g*h_t, g*a_t) \text{ for $\forall g \in G$.}
\end{align}
Moreover, when we already know the next action to take $a_{t+1} = \pi(h_t)$, the next action to take when the input is a transformation of $h_t$ will be a similar transformation of $a_{t+1}$:
\begin{align}\label{eq:equivariant_actor}
    \pi(g*h_t) = g*\pi(h_t) \text{ for $\forall g \in G$.}
\end{align}
Visually illustrated in~\cref{fig:2d-symmetry} with $G=F^{xy}$, the next actions to take given the current histories $(h'^1_T, h'^2_T, h'^3_T)$ are transformed versions of $a_{t+1}$ through $F^{xy}$, i.e., $(a'^1_{t+1}, a'^2_{t+1}, a'^3_{t+1})$.

\subsection{Data Augmentation using Domain Symmetry}
\label{sect:data_aug}

As introduced earlier, there is an opportunity for augmenting the training data (i.e., histories) by leveraging the domain symmetry. Given a transformation element $g \in G$, we first define the following augmentation operation on a history $h_t$:
\begin{align}
    g * h_t \coloneqq (g*o_0, g*a_0, \dots, g*a_{t-1}, g*o_t) \,.
\end{align}
In other words, when transforming a history $h_t$ through $g$, we apply $g$ to every action and observation within $h_t$. We now describe how the transformation affects the actions and observations for a square hole with the transformation $F^{xy}$.

\noindent \textbf{Define $g*o$. } While $F^{xy}$ leaves Z components ($t_z, f_z, \tau_z$) unchanged, their effects on $(t_x, t_y), (f_x, f_y), (\tau_x, \tau_y)$ are similar to the effect of $F^{xy}$ on the point $p_0 = (x,y)$ in~\cref{fig:viz-augment}. 

\noindent \textbf{Define $g*a$. } Similar to $g*o$, $F^{xy}$ does not change $a_z$. In contrast, it transforms $(a_x, a_y)$ like transforming $p_0$.

\subsection{Constructing Auxiliary Losses using Domain Symmetry}
\begin{figure*}
    \centering
    \includegraphics[width=0.9\linewidth]{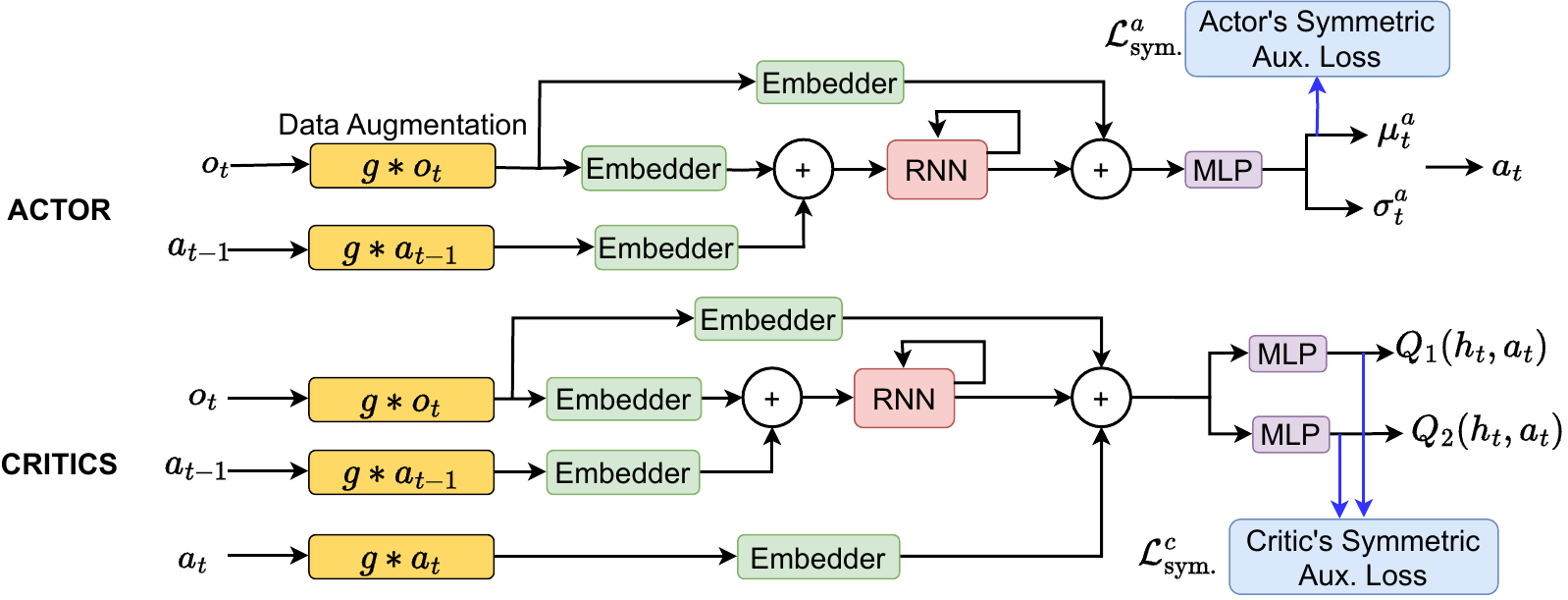}
    \caption{We propose leveraging the domain symmetry through data augmentation (yellow boxes) and symmetric auxiliary losses for the actor and the critic (blue boxes), applied to a recurrent SAC agent~\cite{ni2021recurrent}.}
    \label{fig:method}
    \vspace{-10pt}
\end{figure*}

Now, we show how to enforce the symmetry of an actor-critic agent, as indicated by~\cref{eq:invariant_critic} and~\cref{eq:equivariant_actor}. In this work, we utilize SAC for our base agent due to its sample efficiency, but the method can be easily modified for other actor-critic RL algorithms. As shown in~\cref{fig:method}, our agent is the recurrent SAC agent in~\cite{ni2021recurrent} with embedding layers for encoding observations and actions, recurrent neural networks (RNN) to memorize the embedded features, and multi-layer perceptrons (MLP) to convert the memorized features to actions and Q-values. Our agent, however, has two important differences. First, we perform data augmentation on the input (yellow boxes in~\cref{fig:method}), as described in~\cref{sect:data_aug}. Second, we optimize the two auxiliary losses (blue boxes in~\cref{fig:method}) to force the agent to satisfy~\cref{eq:invariant_critic} and~\cref{eq:equivariant_actor}.

\noindent \textbf{Critic's Symmetric Auxiliary Loss.} As in~\cref{eq:invariant_critic}, we would like the critic to output the same value for $(h, a)$ and $(g*h, g*a)$ inputs for every $g \in G$. This is translated into minimizing the following loss for a critic $Q(h,a)$:
\begin{align}\label{eq:critic_aux_loss}
    \mathcal{L}^c_{\text{sym.}}(Q) = \frac{1}{|G|} \sum_{g \in G}\left( Q(h, a) - Q(g*h, g*a) \right)^2\,.
\end{align}

\noindent \textbf{Actor's Symmetric Auxiliary Loss.} For SAC, whose actor outputs a mean $\mu^a$ and a standard deviation $\sigma^a$, we satisfy~\cref{eq:equivariant_actor} by minimizing the distance between mean outputs (similar to~\cite{nguyen2022leveraging}). Specifically, respectively denoting $\mu^a_h$ and $\mu^a_{g*h}$ to be the means of $\pi(h)$ and $\pi(g*h)$ and defining $g*\mu^a_h$ like $g*a$, we minimize the following loss:
\begin{align}\label{eq:actor_aux_loss}
    \mathcal{L}^a_{\text{sym.}}(\pi) = \frac{1}{|G|}\sum_{g \in G}\left( g*\mu^a_h - \mu^a_{g*h} \right)^2 \,.
\end{align}

\begin{algorithm}
\caption{Symmetry-aware Recurrent SAC}\label{alg:RSAC_Aux}
\begin{algorithmic}[1]
\Require Policy $\pi_\theta$, Q-functions $Q_{\phi_{1, 2}}$ and targets $\bar{Q}_{\bar{\phi}_{1, 2}}$
\Require A group transformation $G$ with size $|G|$
\State Select an action $a \sim \pi_\theta(.|h)$ and execute until done
\State Store the episode in replay buffer $\mathcal{D}$ if it is done
\For{$i \leq \texttt{max. number of updates}$}
    \State Sample $b$ episodes $\{h, a, r, h', d\}$ from $\mathcal{D}$
    \State Augment data to get $b|G|$ episodes: $$(h, a, r, h', d) \leftarrow (g * h, g*a, g*r=r, g*h', g*d=d)$$
    \State Compute targets by sampling $\tilde{a}' \sim \pi_\theta(. | h')$: $$y = r + \gamma (1-d)\big( \min\limits_{i=1, 2} \bar{Q}_{\bar{\phi}_i}(h', \tilde{a}') - \alpha \log \pi_\theta(\tilde{a}' | h')
    \big)$$
    \State Update Q-functions by minimizing:
    $$(Q_{\phi_i} (h, a) - y)^2 + \mathcal{L}^c_{\text{sym.}}(Q_{\phi_i}) \text{ for $i=1, 2$}$$
    \State Update policy by minimizing:
    $$-\min_{i=1, 2} Q_{\phi_i} (h, \tilde{a}_\theta (h)) + \alpha \log \pi_\theta (\tilde{a}_\theta (h) | h) + \mathcal{L}^a_{\text{sym.}}(\pi_\theta) \,,$$
    where $\tilde{a}_\theta(h)$ is a sample from $\pi_\theta(.|h)$.
    \State Update target networks with a coefficient $\rho \in (0, 1)$ $$\bar{\phi}_i \leftarrow \rho \bar{\phi}_i + (1-\rho) \phi_i \text{ for $i=1, 2$}$$
\EndFor
\end{algorithmic}
\end{algorithm}

\noindent \textbf{\cref{alg:RSAC_Aux} }summarizes our approach. The data augmentation is performed in line 5 after a batch of episodes is sampled from the replay buffer. Notice that the transformation $g$ leaves the reward ($r$) and the terminal signal ($d$) unchanged. The two auxiliary losses are optimized when updating the Q-functions (line 7) and the policy (line 8). Notice that there are two Q-functions as required by SAC.

\section{LEARNING IN SIMULATION}
We first perform simulation learning to gauge our method's potential sample efficiency and performance.
\subsection{Peg-In-Hole with a Simulated Soft Wrist}
Our soft wrist is modified from a rigid one in RoboSuite~\cite{robosuite2020}. For simplicity, we use no gripper, instead assuming that the peg is already attached to the arm (see~\cref{fig:coordinates}).

\noindent \textbf{UR5e with a Soft Wrist.} We model a soft wrist using MuJoCo~\cite{todorov2012mujoco}, which can efficiently simulate a spring with a ``tendon". Furthermore, we add a free joint underneath the soft wrist to allow natural movements. Another challenge is that MuJoco does not support detecting contacts involving a non-convex mesh such as a hole. To overcome this, we manually decompose the hole into convex meshes.

\noindent \textbf{Mounting an F/T Sensor.} MuJoCo currently only supports an F/T sensor to measure the forces and torques between two connected bodies. However, the peg and the hole are independent bodies connected by tendons. As a solution, we create a dummy body with a tiny mass \emph{on the peg} and use an F/T sensor to get the signals between the dummy and the peg body. The measured signals should be close to the desired ones due to the small mass of the dummy body.

\noindent \textbf{Demonstrations }are used to overcome the reward sparsity. Humans generate demonstrations by looking at the simulation screen to control the soft wrist through a joystick. Without the F/T feedback, these demonstrations can be considered MDP solutions because visual feedback from the screen can reveal the peg-to-hole pose. However, even though we are trying to learn a POMDP policy, it is possible to learn good POMDP policies using MDP demonstrations when the behavioral overlapping is significant (see~\cite{nguyen2022leveraging, walsman2022impossibly}).

\subsection{Results}

\begin{figure*}[htbp]
    \centering
    \includegraphics[width=\linewidth]{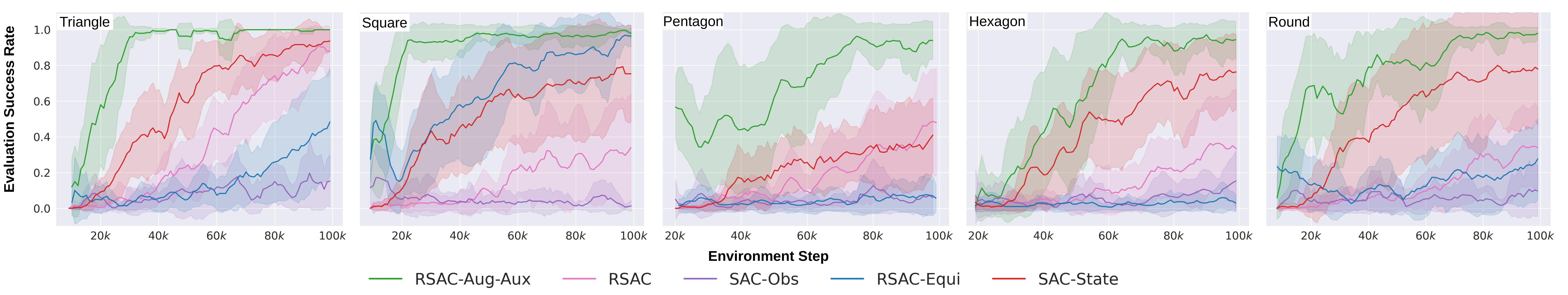}
    \vspace{-15pt}
    \caption{Evaluation success rates averaged over six seeds \emph{in the second training phase} with shaded one standard deviation. \texttt{RSAC-Equi}~\cite{nguyen2023equivariant} and \texttt{RSAC-Aux-Equi} use the specific symmetries listed in the row Best in~\cref{tab:peg_types}. The number of demonstrations used for each domain is listed in the final row of~\cref{tab:peg_types}.}
    \label{fig:main_result}
    \vspace{-15pt}
\end{figure*}

\noindent \textbf{Agents.} \texttt{SAC-State} is trained with the full state and expected as an upper-bound. As a lower-bound, \texttt{SAC-Obs} is trained on only observations. \texttt{RSAC} is recurrent \texttt{SAC-Obs}. These three agents do not leverage data symmetry or data augmentation. Our proposed agent is called \texttt{RSAC-Aug-Aux}. Finally, we compare ours with \texttt{RSAC-Equi}~\cite{nguyen2023equivariant}, which is symmetry-aware by construction. \texttt{RSAC-Equi} outperforms several strong POMDP baselines in a number of pixel-based robot manipulation tasks.

\noindent \textbf{Evaluation Metrics.} We use the \emph{evaluation success rates} to compare the agents. Specifically, during training, the current policy is evaluated on ten episodes for every 1000 training environment steps, and the mean success rates are captured as a single data point. In the simulation, a successful peg insertion is determined by comparing the magnitude of the peg-to-hole pose with a small threshold.

\noindent \textbf{Experimental Conditions.}
We perform learning with five symmetric peg shapes with a clearance of 1 mm, defined as the absolute difference between the radii of the two inscribed circles of the peg and the corresponding hole. The potential transformations and the hole's coordinates for each peg type are summarized in the first two rows of~\cref{tab:peg_types}. The final row of the table shows the number of expert episodes used during training, which consists of two phases. \emph{The first phase} is performed entirely on expert data without interacting with the environment. In \emph{the second phase}, the agent is trained on the mixed data: the expert data and the data generated from interacting with the environment. In both phases, we keep the number of gradient updates equal to the number of environment steps. An episode can last up to 200 timesteps or can be terminated early if the peg is too far from the hole.

\begin{table}[t]
  \centering
  \caption{Additional information about domains.}
  \label{tab:peg_types}
  \begin{tabular}{lccccc}
    \textbf{Shape} & Triangle & Square & Pentagon & Hexagon & Round \\
     & \includegraphics[height=0.7cm]{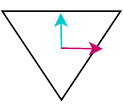} &  \includegraphics[height=0.7cm]{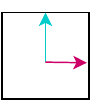} &
    \includegraphics[height=0.7cm]{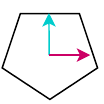} & \includegraphics[height=0.7cm]{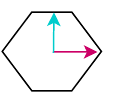} & \includegraphics[height=0.7cm]{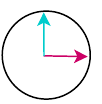}  \\ \midrule
    \textbf{Symm.} &  $\scalemath{0.9}{F^{y},R^3}$ & $\scalemath{0.9}{F^{xy},R^{4}}$ &  $\scalemath{0.9}{F^{y},R^5}$ & $\scalemath{0.9}{F^{xy},R^{6}}$ & $\scalemath{0.9}{F^{xy},R^{4}}$ \\
    \textbf{Best} &  $\scalemath{0.9}{R^3}$ & $\scalemath{0.9}{F^{xy}*R^4}$ &  $\scalemath{0.9}{R^5}$ & $\scalemath{0.9}{R^6}$ & $\scalemath{0.9}{F^{xy}*R^4}$ \\
    \textbf{Demos} & 100 & 100 & 200 & 200 & 100 
  \end{tabular}
\end{table}

\noindent \textbf{Learning Curves.} We illustrate the learning curves of the second training phase averaged over six seeds in~\cref{fig:main_result} (with the specific transformations in the Best row of~\cref{tab:peg_types}), where clearly \texttt{RSAC-Aug-Aux} outperform other baselines. Unexpectedly, it is equal to or better than \texttt{SAC-State} in all domains. However, we expect that if utilizing the symmetry, \texttt{SAC-State} would perform the best. \texttt{RSAC-Equi} performs worse than \texttt{RSAC-Aug-Aux} even in essence, it applies the same idea as ours but in a direct way. A potential reason is the limited exploration capability when strictly enforcing symmetry from the start. Furthermore, training \texttt{RSAC-Equi} takes much more time than training \texttt{RSAC-Aug-Aux}, e.g., 35 v.s. 5 hours for a square peg with the group $F^{xy}*R^4$ and 100k training environment steps. The reason is that, unlike traditional neural networks, equivariant ones~\cite{e2cnn} require additional computation to guarantee the desired symmetry. For POMDPs, even more computation is needed when RNNs\footnote{\cite{nguyen2023equivariant} uses manually constructed RNNs, which are even slower.} might need to process long episodes under a big transformation group. \texttt{SAC-Obs} without memory does not learn in any domain, indicating our domains are true POMDPs. With memory, \texttt{RSAC} improves over \texttt{SAC-Obs}, but its performance is still unsatisfactory in most domains.

\subsection{Ablation Study}
\begin{figure}[htbp]
    \centering
    \includegraphics[width=\linewidth]{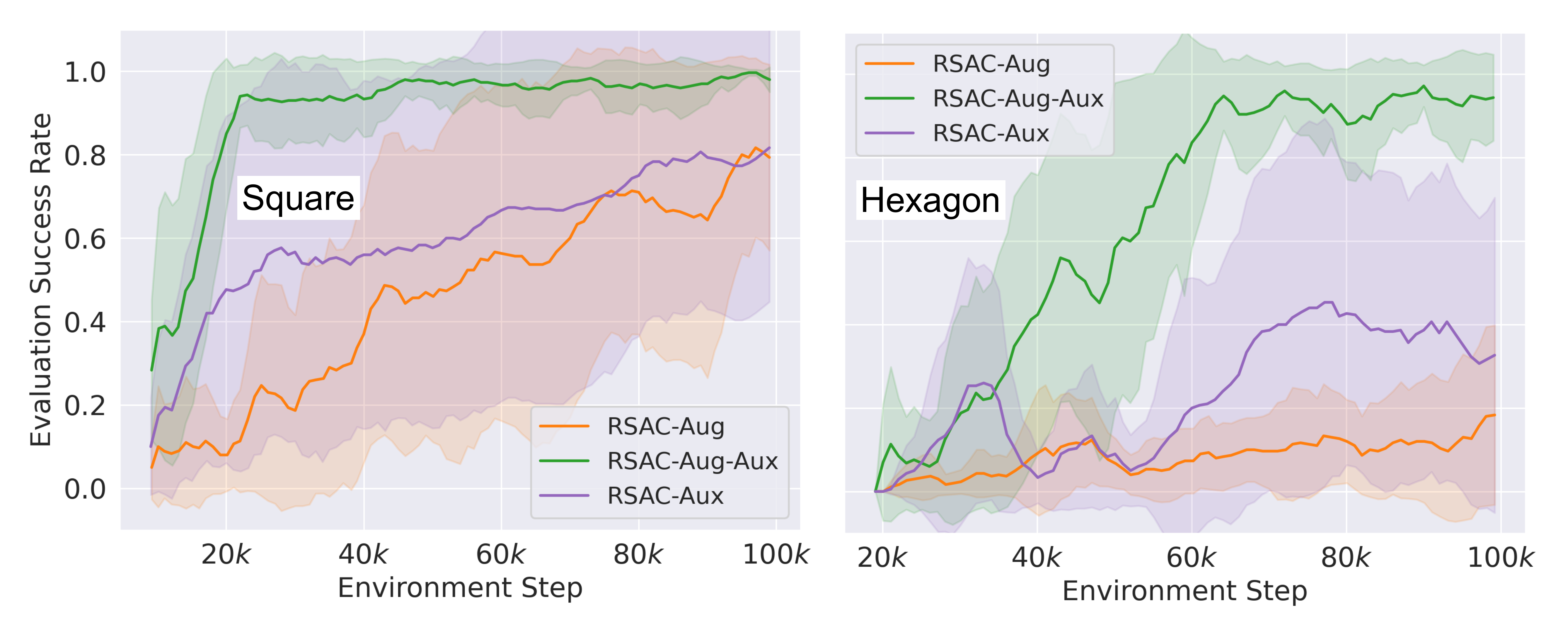}
    \vspace{-15pt}
    \caption{Using either data aug. or aux. losses (six seeds).}
    \label{fig:ablation_result_0}
\end{figure}

\noindent \textbf{Using Either Data Augmentation Or Auxiliary Losses.} For square and hexagonal pegs, in~\cref{fig:ablation_result_0}, we compare \texttt{RSAC-Aug-Aux} with its two variants using either auxiliary losses (\texttt{RSAC-Aux}) or data augmentation (\texttt{RSAC-Aug}). The result shows that combining the two techniques yields much better performance than using only one.

\begin{figure}[htbp]
    \centering
    \includegraphics[width=\linewidth]{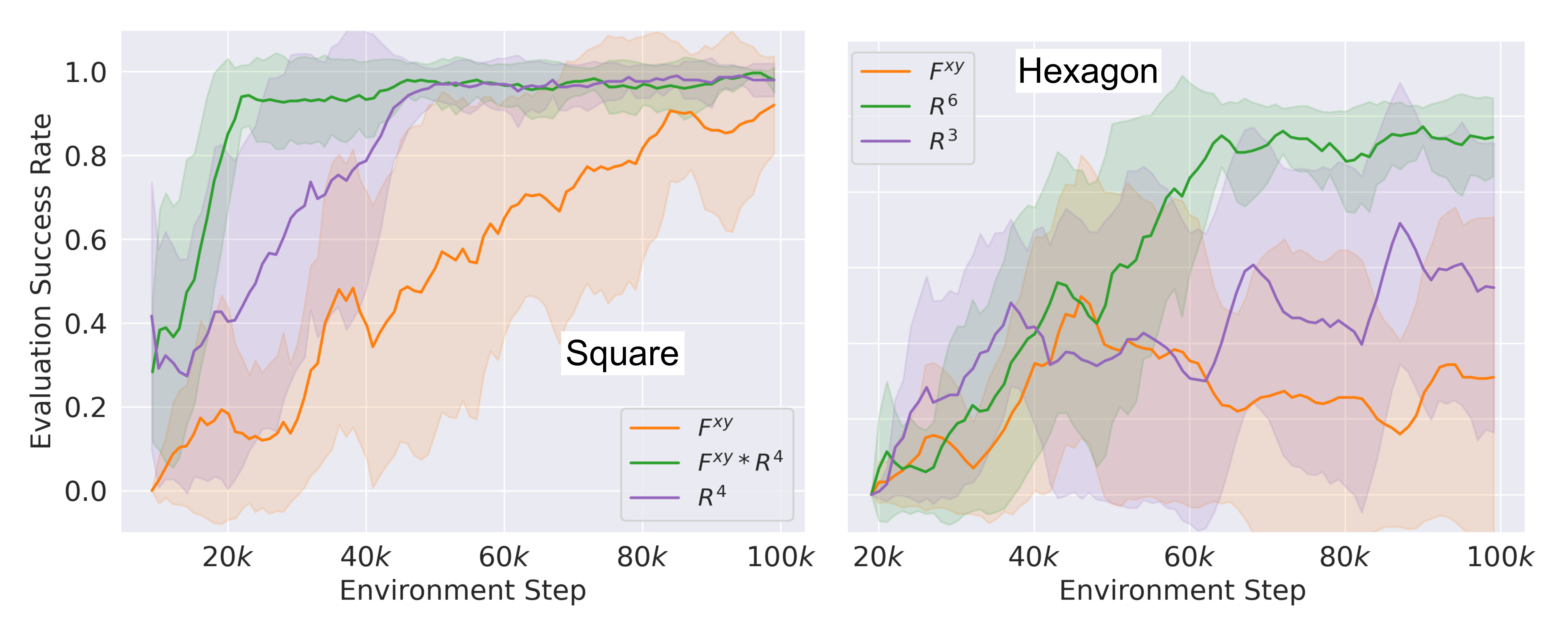}
    \vspace{-15pt}
    \caption{Performance using different symmetries (six seeds).}
    \label{fig:ablation_result_1}
\end{figure}

\noindent \textbf{Leveraging Different Symmetries.} In~\cref{fig:ablation_result_1}, we compare the performance of \texttt{RSAC-Aug-Aux} when leveraging $F^{xy}$, $R^{4}$, and $F^{xy}*R^{4}$ for a square peg. Similarly, we plot the performance with a hexagonal peg when leveraging $F^{xy}$, $R^{3}$, and $R^6$. It can be seen that the performance improves with a bigger group, i.e., $F^{xy} \rightarrow R^4 \rightarrow F^{xy}*R^{4}$. However, the computation will also be more expensive with a bigger batch and group size. Moreover, optimal hyper-parameters will likely change when the batch size drastically changes.

\section{LEARNING DIRECTLY ON HARDWARE}
\subsection{Setup}

\begin{wrapfigure}[13]{R}{0.4\linewidth}
  \centering
  \includegraphics[width=1.0\linewidth]{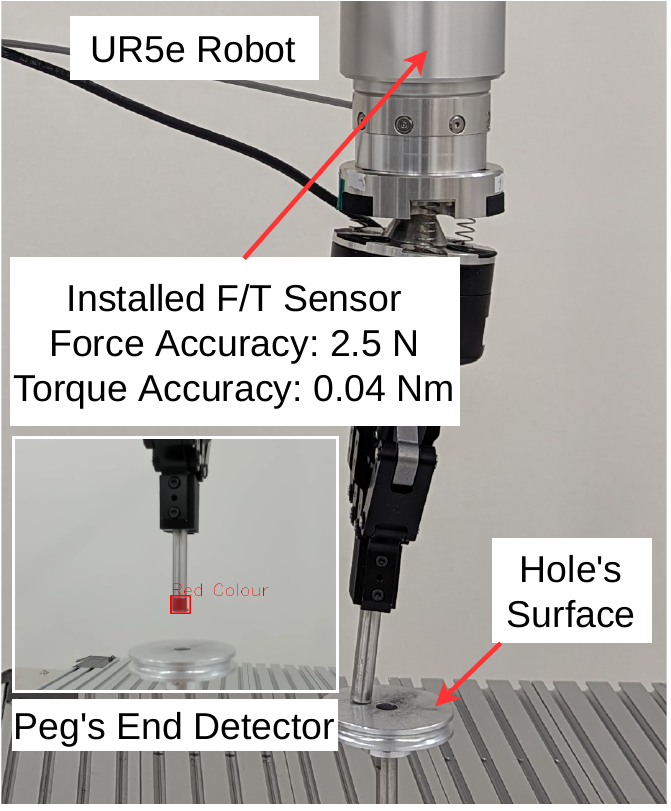}
  \vspace{-14pt}
  \caption{Robot Setup.}
  \label{fig:robot_setup}
\end{wrapfigure}

We perform learning on hardware, using a soft wrist~\cite{von2020compact} mounted on a UR5e robot (see~\cref{fig:robot_setup}) and a round and a square peg, each with a 1-mm clearance. The F/T sensor installed in UR5e (parameters in~\cref{fig:robot_setup}) is located before the soft wrist, which measures the F/T signals applied to the entire arm. This differs from the simulation setting, in which only the F/T signals applied to the peg are measured. The peg's initial position is randomized to touch the hole's surface to prevent mid-air vibration when moving. We filter the signals with a low-pass filter (50-Hz cut-off frequency and 500-Hz sampling frequency). We also remove the F/T offsets when starting an episode due to sensor drift.

\noindent \textbf{Experimental Conditions. } We contrast the performance of \texttt{RSAC-Aug-Aux} ($F^{xy}*R^4$) and \texttt{RSAC}. Episodes are limited to 50 timesteps, with each agent receiving 100 expert episodes, corresponding to 30 minutes for data collection. A policy evaluation of five episodes is performed every 500 steps with $\delta_x, \delta_y \in [-0.01, 0.01]$ and $\delta_z \in [-0.0025, 0.0025]$ m. To promote gentle force application, we impose a -0.1 penalty for forces beyond the [0, 10] N range and a -5.0 penalty for exceeding a certain peg-hole distance. Task successes rely on comparing $t_{xyz}$ with a small threshold and the peg's end detector binary output (see~\cref{fig:robot_setup}). The end is marked red, and the detector returns false on successful insertions. The detector helps filter out false positives when the peg tilts inward the center with very small $t_{xyz}$, but no insertion is performed.

\subsection{Results}
\begin{figure}[htbp]
    \centering
    \includegraphics[width=\linewidth]{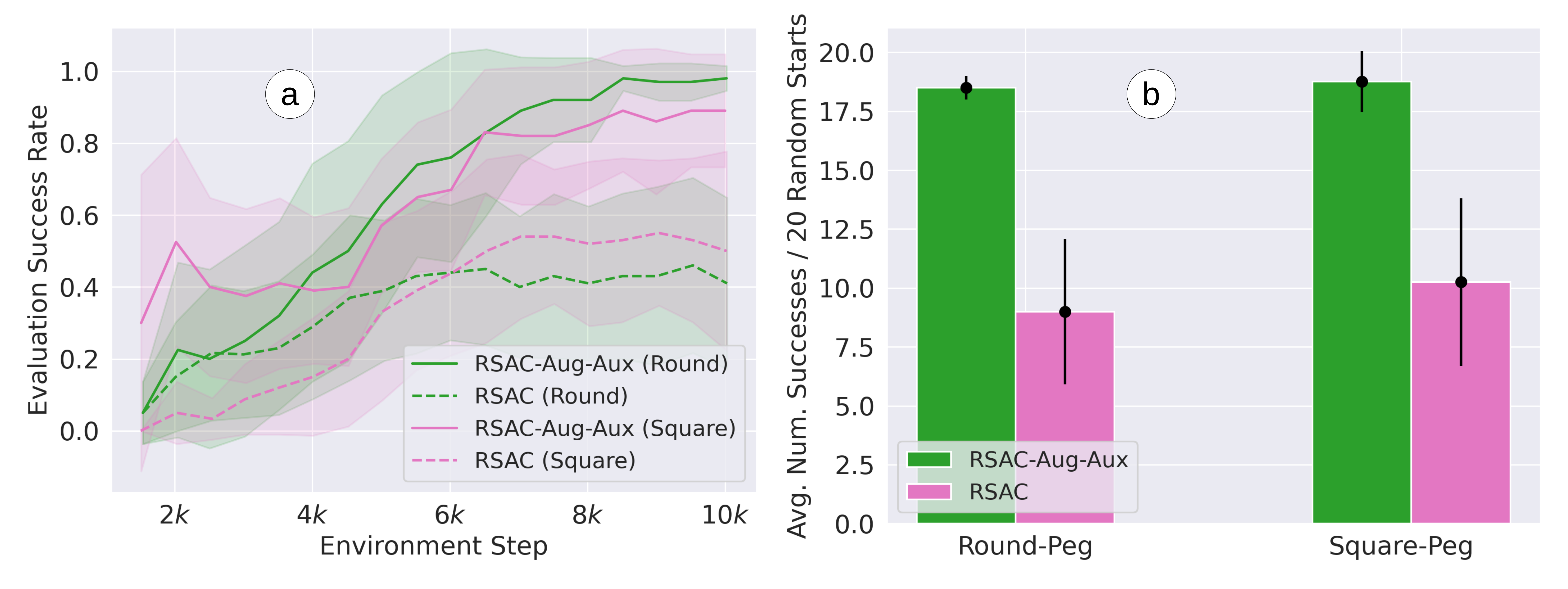}
    \caption{Real-robot results (four seeds): a) Learning curves, b) Number of successes out of 20 runs of the final policies.}
    \label{fig:real_result}
\end{figure}

We perform training in 10k environment steps, which take about 3 hours to finish. \cref{fig:real_result}a shows that \texttt{RSAC-Aug-Aux} significantly outperforms \texttt{RSAC} regarding the validation success rates. Moreover, the final policies of \texttt{RSAC-Aug-Aux} are also more successful when extensively evaluated on 20 different random starting positions (see \cref{fig:real_result}b). During training, compared to \texttt{RSAC-Aug-Aux}, \texttt{RSAC} wastes more time exploring randomly, e.g., moving outside of the workspace or randomly moving on the hole's surface instead of heading towards the center where the hole is.

\section{Discussion}

\noindent \textbf{Simulation: Surpassing Demonstrators.} The learned policies are better than the demonstrations. Because human demonstrators rely entirely on visual feedback, they first push down the gripper to touch the hole to determine the relative position between the peg and the hole before moving into it. In contrast, the learned policy directly goes to the center of the hole (relying on $t_{xyz}$) in the air, fitting the peg into the hole, adjusted if necessary using the F/T feedback. This approach consistently finishes the task much quicker 
 (see~\cref{tab:length_compare} and video). Qualitatively, the learned policies exhibit greater proficiency in combining all three action axes.

\begin{table}[t]
  \centering
  \caption{\texttt{RSAC-Aug-Aux} v.s. Demos: Average number of steps for task completion in simulation (\emph{smaller is better}).}
  \label{tab:length_compare}
  \begin{tabular}{lccccc}
    Peg Shape &  $\triangle$ &  $\square$ & \pentago & \hexago & \Circle \\ \midrule
    Converged \texttt{RSAC-Aug-Aux} &  20.3 & 14.4 &  29.4 & 21.3 & 22.7 \\
    Human Demonstrations & 90.4 & 86.6 & 100.7 & 92.8 & 78.8 
  \end{tabular}
  \vspace{-10pt}
\end{table}

\noindent \textbf{Simulation: Policy Generalization to Peg Shapes.}
Given that Z-axis rotations are unnecessary, we expect a learned policy in one shape to work with another peg shape. Indeed, the policy learned with a triangular peg nearly succeeds 100\% with a pentagonal or hexagonal peg (see our video). The same policy, however, does not transfer to a square or round peg because these domains are initialized differently.

\noindent \textbf{Learning on Hardware: Policy Analysis \& Generalization.} During training, mastering the skill of applying a small force to slide the peg over the hole's surface proves most challenging. Our agent learns to use slight vibrations and movements for careful hole-searching near the center (see our video), representing the information-gathering actions of a POMDP solution. In contrast, the demonstration strategy is an MDP solution, tilting the peg and moving purposefully toward the hole's known position with larger movements. Intriguingly, our agent's learned policies remain robust against manual disruptions and partly transfer to a round peg with 20 times smaller clearance (50 $\mu$m instead of 1 mm), succeeding in 9/20 attempts. Furthermore, similar to the shape generalization in simulation, the learned policy for a metal round peg achieves a 20/20 success rate with a 3D-printed square peg with the same clearance of 1 mm.

\section{CONCLUSIONS}
This study enhances a recurrent SAC agent for solving symmetric POMDPs by integrating data augmentation and auxiliary losses. 
Our method's versatility extends beyond the peg-in-hole task and can be adapted for other scenarios.
A limitation of our method is the requirement of near-perfect symmetries for optimal performance. Yet, minor imperfections in real-robot experiments, like surface unevenness or inconsistency, did not significantly affect the performance. Also, a recent study shows the benefits of leveraging symmetries in imperfect settings~\cite{wang2022surprising} and the improved robustness informing of the contact states using tactile sensors~\cite{royo2023learning}, which can potentially apply to our POMDP context.




\clearpage
\bibliographystyle{IEEEtran}
\bibliography{refs}

\end{document}